\documentclass[10pt,twocolumn,letterpaper]{article}

\usepackage{iccv}
\usepackage{times}
\usepackage{epsfig}
\usepackage{graphicx}
\usepackage{amsmath}
\usepackage{amssymb}
\usepackage{algorithm}
\usepackage{algorithmic}
\usepackage{bbm}
\usepackage{amsfonts}
\usepackage{mathrsfs}
\usepackage{multirow}
\usepackage{booktabs}

\usepackage[pagebackref=true,breaklinks=true,letterpaper=true,colorlinks,bookmarks=false]{hyperref}

\iccvfinalcopy 

\def\iccvPaperID{406} 

\begin{document}

\title{IntraLoss: Further Margin via Gradient-Enhancing Term for Deep Face Recognition}

\author{
Chengzhi Jiang{},~ Yanzhou Su{},~ Wen Wang{},~ Haiwei Bai{},~ Haijun Liu{},~ Jian Cheng{$^{*}$}\\
\vspace{-8pt}{\small~}\\
}

\maketitle
\thispagestyle{empty}

\let\thefootnote\relax\footnotetext{$^{*}$Corresponding Author}

\begin{abstract}
Existing classification-based face recognition methods \cite{deng2019arcface, wang2018additive, wang2018cosface, liu2017sphereface} have achieved remarkable progress, introducing large margin into hypersphere manifold to learn discriminative facial representations. However, the feature distribution is ignored. Poor feature distribution will wipe out the performance improvement brought about by margin scheme. Recent studies \cite{zhao2019regularface, duan2019uniformface} focus on the unbalanced inter-class distribution and form a equidistributed feature representations by penalizing the angle between identity and its nearest neighbor. But the problem is more than that, we also found the anisotropy of intra-class distribution. In this paper, we propose the `gradient-enhancing term' that concentrates on the distribution characteristics within the class. This method, named $ \emph{IntraLoss} $, explicitly performs gradient enhancement in the anisotropic region so that the intra-class distribution continues to shrink, resulting in isotropic and more compact intra-class distribution and further margin between identities. The experimental results on LFW, YTF and CFP-FP show that our $ \emph{IntraLoss} $ outperforms state-of-the-art methods by gradient enhancement, demonstrating the superiorty of our method. In addition, our method has intuitive geometric interpretation and can be easily combined with existing methods to solve the previously ignored problems.
\end{abstract}

\section{Introduction}
\label{sec:intro}

For the past few years, Deep Convolutional Neural Networks has greatly boost-ed state-of-the-art in Face Recognition, which makes deep CNN a primary way for the problem. Face Recognition, as one of the most common computer vision task, consists of two sub-tasks: face identification and face verification. Face identification, a one-to-many task, assigns a known identity to a given facial image from database. And face verification, a one-to-one task, determines whether a pair of facial images comes from the same identity. For testing protocols, according to the consistency of identity of training set and test set, it can be divided into two kinds: 1) Open-set protocol, where testing identities may not come from training set, and 2) Close-set protocl, where training set and testing set have the same identities. Face Recognition can be regarded as classification tasks or metric learning tasks. Under the premise of close-set protocol, face recognition is viewed as a classification task, and the face identity is directly output via softmax layer. While from the perspective of open-set, face recognition is more like a metric learning task, and the corresponding identity is obtained by performing feature comparison with the identity in the database. Under such a circumstance, feature extractor requires more discriminative capabilities, which means that features of the same class are as compact as possible(intra-class compactness), whereas features of different class are far enough apart(inter-class separability). It is obvious that open-set face recognition is closer to real-world use and applications, yet with more challenge. 

A standard deep CNN-based face recognition pipline consists of four stages: face detection, face alignment, feature representation, and feature comparison. Early deep learning-based studies \cite{taigman2014deepface,huang2008labeled,sun2014deep,yi2014learning}, in the manner of multi-class classification problem, build deep convolutional neural networks to extract high-level facial features, used the softmax loss to supervise training procedure, and improve performance through elaborate network design and large scale dateset.
Deepface \cite{taigman2014deepface} proposes a local convolution layer to learn different convolutional kernels for different positions on the feature map to adapt to different features in different regions of the face. DeepID \cite{sun2014deep} constructs a network with the last hidden layer connected to the third and fourth convolutional layers, and the dimension of the last hidden layer is fixed to 160, which is much smaller than the number of identities to form a compact and predictive feature representation. Deep CNNs learn from the data, and the scale of the data greatly affects the generalization performance. Recent years have presented several large-scale face datasets, such as CASIA-WebFace \cite{yi2014learning}, LFW \cite{huang2008labeled}, MS-Celeb-1M \cite{guo2016ms}, Youtube Face(YTF) \cite{wolf2011face}, VGGFace \cite{parkhi2015deep}. More than that, data synthesis method \cite{masi2016we} has also been published, further inreasing the amount of the data, improving the generalization performance and preventing overfitting. 


Apart from the two factors above, loss function plays a ciritical role in determining the discriminative capability of learned facial representations. Through softmax loss, which is widely used to train deep CNNs, only separable feature representations can be learned, with limited discriminative capacity. For face recogniton, feature representations should be not only separable, but also discriminative. To motivate better discriminative feature performance, \textbf{metric-based} loss was introduced.  Contrastive loss \cite{chopra2005learning} and triplet loss \cite{schroff2015facenet} learn more efficient features directly from training pairs and triplets, leading to Euclidean margin(the largest intra-class distance is smaller than the shortest inter-class distance in Euclidean space). Nevertheless, the careful selection of pairs and triplets is very difficult and time-consuming, and it will inevitably significantly increase the computational complexity and cause slow convergence and instability. 

In contrast, softmax loss has simple sampling and fast convergence, hence some works \cite{sun2014deepid2, wen2016discriminative} combine softmax loss and metric loss together to train CNNs. Furthermore, many \textbf{classification-based} research studies \cite{liu2016large, liu2017sphereface, ranjan2017l2, wang2017normface, wang2018cosface, wang2018additive, deng2019arcface} has been proposed. \cite{ranjan2017l2, wang2017normface} normallize features and weights parameters and map them onto hypersphere manifold, and show that the normalization operation can boost the final performance. Based on this, SphereFace \cite{liu2017sphereface} imposes angular margin to hypersphere in a multiplicative way. More than that, AM-softmax \cite{wang2018additive, wang2018cosface} and ArcFace \cite{deng2019arcface} respectively improved softmax loss in the form of additive cosine margin and additive angular margin. 

However, the above methods all focus on introducing margin into loss function to enhance the discriminative capacity, but neglect to consider the overall feature distribution on the hypersphere. Latest researches, such as RegularFace \cite{zhao2019regularface} and UniformFace \cite{duan2019uniformface} , point out that the inter-class distribution is highly nonuniform, which leads to less feature discrimination despite the large intra-class margin, and new objective functions are proposed to learn equidistributed feature representation. Beyond that, when we look at the intra-class distribution, we found another problem that the shape of each intra-class distribution is an anisotropic polygon, which is a system defect of all classification-based methods \textbf{due to the anisotropy of the feature space partition when classifying}. 
Geometrically speaking, when margin works, the origin classification decision boundaries are pushed closer to the center of each class. Accordingly, the intra-class distributiion becomes compacter, but the original distribution characteristics are maintained. \textbf{The anisotropy makes, in different orientation, the same distance from class center brings about different gradients.} We suggest that such a phenomenon will cause incomplete optimization of intra-calss distribution, and compromise the inter-class margin. 


In this work, we focus on the other side of feature distribution: intra-class distribution, and propose a gradient-enhancing term, named \textbf{\emph{IntraLoss}}, to impose a novel supervising signal upon hypersphere. This gradient-enhancing term explicitly restores gradient where gradient disappears farther away from the class center, resuming optimizing process. Consequently, the intra-class variations is further reduced, resulting in further margin on the hypersphere. 

Our contributions are summarized as follows: 
\begin{itemize}
	\item First, we raise and discuss the problem of \textbf{nonequivalent} intra-class gradient distributions in different orientation on hypersphere manifold, and the accompanying problem of class center diviation. 
	\item Second, we propose a novel gradient-enhancing term, called \textbf{IntraLoss}, which explicitly narrows intra-class variations to encourage further margin by restoring gradient. To be best of our knowledge, this is the first attempt to enhance feature discrimination by using such a loss objective on the hypersphere manifold.
	\item Third, our proposed gradient enhancing method is orthogonal with, and can be easily implemented into existing methods, and it is adaptive according to the distributiion of each class on the hypersphere with no extra hyperparameters introduced.
	\item And last, we present extensive experiments on LFW \cite{huang2008labeled}, Youtube face(YTF) \cite{wolf2011face} and CFP-FP \cite{cfp-paper}, and our method outperforms most of the existing methods to above datasets.
\end{itemize}

\begin{figure*}  
	\centering
	\includegraphics[width=14cm]{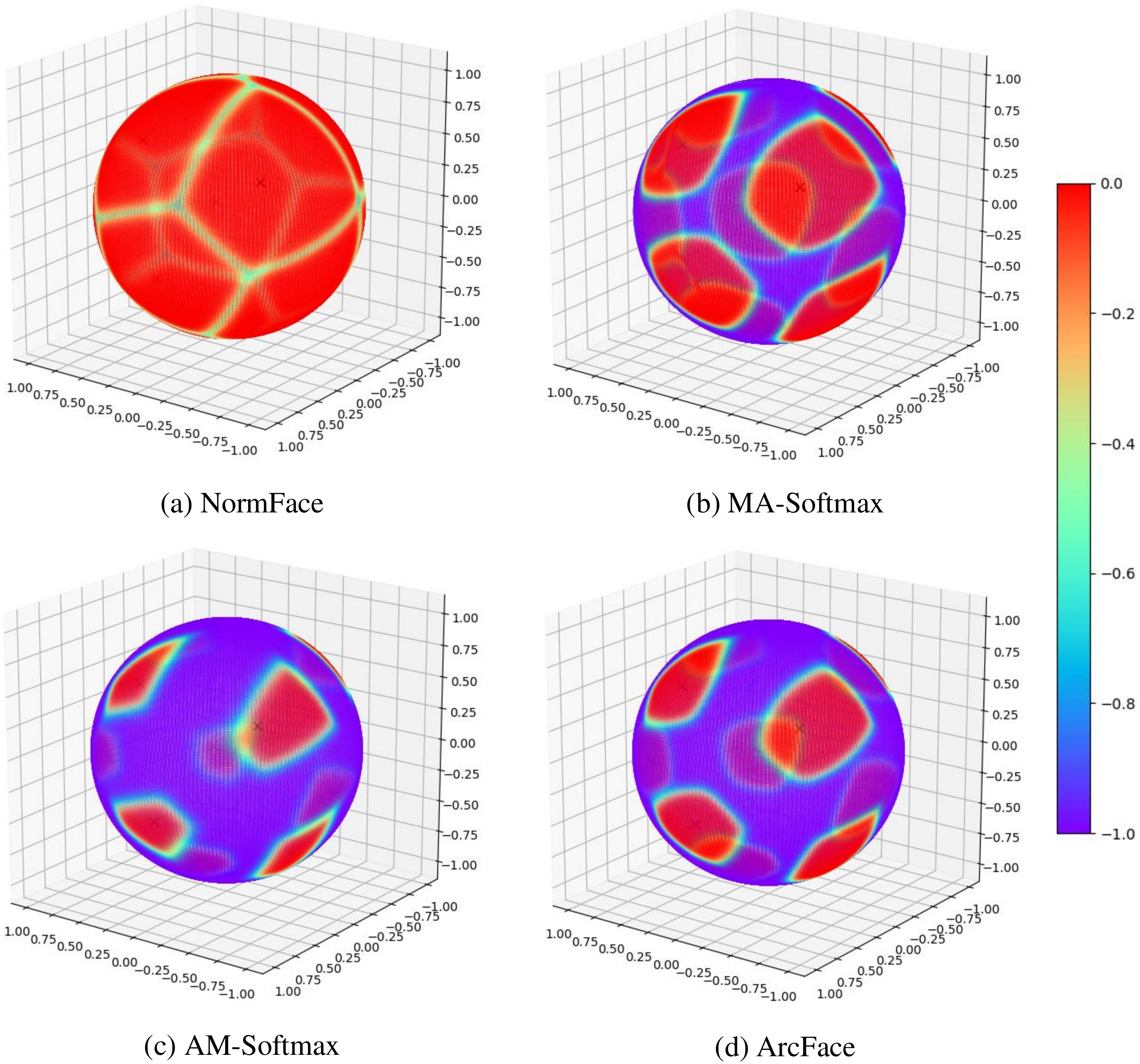}
	\caption{Illustration of the anisotropy intra-class distribution. From the figure, we can observe that: 1) \emph{L2}-normalization maps feature onto a fixed radius hypersphere, but without the margin penalty, the learned features are still separable. 2) The classification loss partition feature space in an anisotropic way. 3) When margin parameter works, the distance between classes increases and the distribution of each class becomes more compact. 4) Different types of margin schemes result in different forms of margin, but the intra-class distribution has the same anisotropy problem. }
	\label{fig:distribution_illus}
\end{figure*}

\section{Related work}
\label{sec:relatedwork}
To our best knowledge, based on the large margin loss functions in use, there are two major types methods: metric-based methods and classification-based methods.

\subsection{Metric-Based Methods}
The purpose of Metric-Based Methods are to obtain a similarity(semantic distance) function in Euclidean space through learning, reduce the distance between homogeneous samples, and increase the distance between different class of samples. For open-set face recognition,  metric between samples directly determines the final result.

\textbf{Metric Loss.} \ Since the open-set face recognition is similar to metric learning, some works directly implement metric loss to Networks. Contrastive Loss \cite{chopra2005learning} constructs a Siamese Networks to perform representation learning, and two facial images are fed into networks to obtain their respective feature representations. By penalizing the distance between image pairs, make the images of the same class as close as possible, and images of different classes are far enough. Moreover, Triplet Loss \cite{schroff2015facenet} accepts three facial images at once, two of which are from the same class (the anchor and positive) and the other from a different class (the negative). Triplet loss minimizes the distance between the same class, and maximizes the distance between different classes, bringing out that the largest intra-class distance is less than the smallest inter-class distance.

\textbf{Joint Loss.} \ Note that metric loss is difficult to converge, and needs careful and time-consuming sampling strategy. DeepID2 \cite{sun2014deepid2} forms a joint loss function based on DeepID's network \cite{sun2014deep}, which greatly improves performance. Concretely speaking, the original convolutional neural network used softmax loss as objective function, so-called identification signal. In DeepID2 \cite{sun2014deepid2}, a verification signal, known as Contrastive loss, is introduced. The two signals are combined using a weighted manner. In regularization way, Center Loss \cite{wen2016discriminative} penalizes the distance between feature and the corresponding center to reduce the intra-class variations, and finally supervises the training together with sofmax loss.

\subsection{Classification-based Methods}
Classification-based Methods are designed to take advantage of simple sampling of the classification loss and fast convergence, and to enhance the discrimination of feature representation by formulating a large-margin variant. Compared with Euclidean metric based methods, classification-based methods remaps feature space onto a hypersphere manifold and measure similarity between samples by cosine or angular distance. 

\textbf{Normalization.} \ In practical training, the inner product is the most widely-used similarity metric in last fully connect layer. While in testing phase, cosine similarity is performed, which is different from used in training phase. To align the metrics during the training and testing phases, NormFace \cite{wang2017normface} builds a cosine layer by $ \emph{L2} $-normalizing the weight parameters and the input features. And Scale parameter is used to control cosine value to help the algorithm converge. As a result, normalization advances the accuracies of state-of-the-art on LFW and YTF.

\textbf{Multiplicative Margin.} \ After decoupling the magnitude and angle of the inner-product layer, L-Softmax \cite{liu2016large} firstly incorporates an angular margin with decoupled Softmax loss, multiplying with the angle between feature and the class agent of related class. L-Softmax only looks at margin and does not normalize weights or features. On this basis, SphereFace \cite{liu2017sphereface} not only imposes a multiplicative angular margin into loss function, but also normalizes the weight parameters, making the new method more geometrically interpretable and better than former works.

\textbf{Additive Margin.} \ NormFace \cite{wang2017normface} shows the effectiveness of normalizing both weight parameters and features, and further efforts are made in the subsequent works. AM-Softmax \cite{wang2018additive} incorporates an additive margin with cosine value in the form of \(cos\theta - m\) , and also emphasizes and investigates the importance of feature normalization. Meanwhile, ArcFace \cite{deng2019arcface} imports margin directly to the angle in the same additive way, form like \(cos(\theta + m)\). Extensive experiments on several authoritative benchmarks prove that ArcFace outperforms most of the existing Classification-based Methods.

\begin{table*}
	\small
	\caption{ Decision boundary of different class in binary situation demonstrate how boundaries change from common to separate. Notice that, $ \theta_{j} $ is the angle between $ W_j $ and $ \textbf{\emph{f}} $. }
	\label{tab:decision_boundaries}
	\centering
	\setlength{\tabcolsep}{0.5cm}
	\begin{tabular}{l|c}
		\toprule[2pt]
		Margin Scheme  & Decision Boundarys \\
		\toprule[1pt]
		\toprule[1pt]
		without margin scheme & $ \cos \theta_1 - \cos \theta_2 = 0 $ \\
		\hline
		\multirow{2}{*}{ Multiplicative Angular Margin }  & \multirow{1}{*} {$ \cos m\theta_1 - \cos \theta_2 = 0 $ \quad for class 1} \\
		& \multirow{1}{*} {$ \cos \theta_1 - \cos m\theta_2 = 0 $ \quad for class 2} \\
		\hline
		\multirow{2}{*}{ Additive Cosine Margin }  & \multirow{1}{*} {$ (\cos \theta_1 - m) - \cos \theta_2  = 0 $ \quad for class 1} \\
		& \multirow{1}{*} {$ \cos \theta_1 - (\cos \theta_2 - m) = 0 $ \quad for class 2} \\
		\hline
		\multirow{2}{*}{ Additive Angular Margin } & \multirow{1}{*} {$ \cos (\theta_1 + m) - \cos \theta_2 = 0 $ \quad for class 1} \\
		& \multirow{1}{*} {$ \cos \theta_1 - \cos (\theta_2 + m) = 0 $ \quad for class 2} \\
		\toprule[2pt]
	\end{tabular}
\end{table*}

\section{Observation and Motivation}
\label{sec:observation_and_motivation}
In this section, we detail the anisotropic problem. To better understant the causes, in Section \ref{sec:Begin_with_Softmax}, we will first revisit the softmax loss and its variants. And in Section \ref{sec:a_toy_example}, we use a toy example to intuitively show the anisotropic distribution. Then Section \ref{sec:comparison} compares and discusses the different manifestations of the anisotropic intra-class distribution between additive and multiplicative methods. Finally, Section \ref{sec:motivation} states how we are inspired to form such a loss funtion. 

\subsection{Begin with Softmax}
\label{sec:Begin_with_Softmax}
In the framework of classification-based large margin methods, loss function is defined as the combination of a cross-entropy loss, a softmax function and zero-bias inner-product layer. 
Then the original softmax loss can be written as
\begin{align} \label{formu:softmax}
L_{\textbf{\emph{S}}} 
&= -\frac{1}{n}\sum_{i=1}^{n}log\frac{{e^{W^T_{y_i} \hspace{1pt} \textbf{\emph{f}}_i}}}{\sum_{j=1}^{c}e^{{W^T_j}{\textbf{\emph{f}}_i}}} 
\notag
\\
&=-\frac{1}{n}\sum_{i=1}^{n}log\frac{e^{{\lVert W_{y_i} \rVert}{\lVert \textbf{\emph{f}}_i \rVert}{\cos{(\theta_{y_i})}}}}{\sum_{j=1}^{c}e^{{\lVert W_{j} \rVert}{\lVert \textbf{\emph{f}}_i \rVert}{\cos{(\theta_{j})}}}} ,
\end{align}
where $ \textbf{\emph{f}}_i \in \mathbb{R}^{d} $ denotes the $ i $-th input feature of the last fully connected (FC) layer, belonging to $ \emph{y}_i $-th class. $ W_{j} \in \mathbb{R}^{d} $ is the $ j $-th column of the last FC weights $ W \in \mathbb{R}^{d \times c} $. $ d $ denotes the feature dimension, and the batch-size and the number of class is $ n $ and $ c $, respectively. Notice that $ W^T_{y_i} \hspace{2pt} \textbf{\emph{f}}_i = {\lVert W_{y_i} \rVert}{\lVert \textbf{\emph{f}}_i \rVert}{\cos{(\theta_{y_i})}} $ , we get decoupled version of origin on the second line, in which $ \theta_{j} $ is the angle between feature $ \textbf{\emph{f}}_i $ and $ W_{j} $. It is necessary to mention that $ z_{y_i} =  W^T_{y_i} \hspace{2pt} \textbf{\emph{f}}_i $ is also called \textbf{\emph{target logit}} of $y_i$-th class, and this name is used in this article. 
For sample $ \textbf{\emph{f}} $ of class $ y_i $, Softmax loss encourages $ W^T_{y_i} \hspace{2pt} \textbf{\emph{f}} > W^T_{j} \hspace{2pt} \textbf{\emph{f}} \hspace{5pt} \left(i.e. \hspace{5pt} {\lVert W_{y_i} \rVert}{\lVert \textbf{\emph{f}} \rVert}{\cos{(\theta_{y_i})}} > {\lVert W_{j} \rVert}{\lVert \textbf{\emph{f}} \rVert}{\cos{(\theta_{j})}} \right) $ to classify correctly. 
And the decision surface is defined by 
\begin{equation} \label{formu:softmax_surface}
{\lVert W_{y_i} \rVert}{\lVert \textbf{\emph{f}} \rVert}{\cos{(\theta_{y_i})}} = {\lVert W_{j} \rVert}{\lVert \textbf{\emph{f}} \rVert}{\cos{(\theta_{j})}}.
\end{equation}

From a macro perspective, softmax loss deals with multi-classification problem, but near the class boundary, it is more like a linear binary classification problem. 

SphereFace \cite{liu2017sphereface} proposed A-Softmax loss, which constrains $ \lVert W_j \rVert = 1 $ by \emph{L2}-normalization, and angular margin \emph{m} is introduced by multiplying angle $ \theta_{y_i} $, 
\begin{equation} \label{formu:a-softmax}
L_{AS} = -\frac{1}{n}\sum_{i=1}^{n}log\frac{e^{{\lVert \textbf{\emph{f}}_i \rVert}\cdot{\cos{(m_1 \theta_{y_i})}}}}{e^{{\lVert \textbf{\emph{f}}_i \rVert}\cdot{\cos{(m_1 \theta_{y_i})}}} + \sum_{j=1, j \neq y_i}^{c}e^{{\lVert \textbf{\emph{f}}_i \rVert}\cdot{\cos{(\theta_{j})}}}} , 
\end{equation}
where $ m_1 (\ge 1) $ is a hyperparameter that needs to be set manually, and the surface equation becomes 
\begin{equation} \label{formu:a-softmax_surface}
{\lVert \textbf{\emph{f}}_i \rVert}\cdot{\cos{(m_1 \theta_{y_i})}} = {\lVert \textbf{\emph{f}}_i \rVert}\cdot{\cos{(\theta_{j})}}, 
\end{equation}
decision surface is only related to the angle between feature and weight. Only when $ \theta_{y_i} \leq \frac{\theta_{j}}{m} $ can $ \textbf{\emph{f}}_i $ be correctly classified. Under such a condition, the $ y_i $-th intra-class distribution is compressed.

$ \lVert \textbf{\emph{f}}_i \rVert $ varies from sample to sample. Note that feature normalization is widely used as a trick in the test phase of face recognition. Normface \cite{wang2017normface} normalizes both weight and feature during the training phase, mapping them onto a hypersphere of a fixed radius. And scale factor $ s $ is introduced to promote algorithm convergence. Finally the loss function is defined as
\begin{align} \label{formu:normface}
L_{NormS} &= -\frac{1}{n} \sum_{i=1}^{n} log \frac{e^{s \cdot \tilde{W}^{T}_{y_i} \tilde{f}_i }}{\sum_{j=1}^{c} e^{s \cdot \tilde{W}^{T}_{j} \tilde{f}_i }}
\notag
\\
&= -\frac{1}{n} \sum_{i=1}^{n} log \frac{e^{s\cdot \cos{(\theta_{y_i})}}}{\sum_{j=1}^{c} e^{s \cdot \cos{(\theta_{j})}}}, 
\end{align}
where $ \tilde{x} = \frac{ x }{\lVert x \rVert}$ , and $ s $ is a hyperparameter determining the lower bound of loss. On the hypersphere, $ \tilde{W}_j $ is usually regarded as the cluster center of $ j $-th class, spontaneously learned from the training data. So the geodesic distance between the sample and the $ j $-th class center on the hypersphere can be represented by $ \cos\theta_{j} $ (i.e. cosine metric), it's a linear one-to-one mapping.

The $ \emph{L2} $-normalization operation is easy to plug into existing methods using modern deep learning frameworks and can boost the performance of face recognition greatly.  With such a improvement, the A-Softmax can be reformulated as Modified A-Softmax (MA-Softmax), 
\begin{equation} \label{formu:ma-softmax}
L_{MAS} =  -\frac{1}{n} \sum_{i=1}^{n} log \frac{e^{s\cdot \cos{(m_1 \theta_{y_i})}}}{e^{s\cdot \cos{(m_1 \theta_{y_i})}} + \sum_{j=1, j \neq y_i}^{c} e^{s \cdot \cos{(\theta_{j})}}}, 
\end{equation}
and the surface equation is rewritten as follows, 
\begin{equation} \label{formu:ma-softmax_surface}
s\cdot{\cos{(m_1 \theta_{y_i})}} = s\cdot{\cos{(\theta_{j})}}, 
\end{equation}

In multiplicative margin situation, $ \theta_{y_i} $ should be in the range of $ \left[ 0, \frac{\pi}{m_1} \right] $ due to the non-monotonicity of cosine function. To eliminate this limitation and make it optimizable in the network, SphereFace replaces $ \cos(m_1 \theta_{y_i}) $ with a monotonically decreasing funcion, 
\begin{align} \label{formu: monotonic_func}
\psi(\theta_{y_i}) = \frac{(-1)^k \cos(m_1 \theta_{y_i}) - 2k + \lambda \cos(\theta_{y_i})}{1 + \lambda} &,
\notag
\\
\theta_{y_i} \in \left[ \frac{k\pi}{m_1}, \frac{(k+1)\pi}{m_1} \right], k \in \left[ 0, m_1 -1 \right] &, 
\end{align}
during implementation. $ \lambda $ is an additional annealing parameter, which is set to 1000 in the initial stage of training and gradually reduced to a small value. 

Conversely, additive margin methods don't require analogous tricky training strategies. AM-Softmax \cite{wang2018additive, wang2018cosface} imposes margin to cosine value instead of angular in the form as
\begin{align} \label{formu:am-softmax}
L_{AMS} &= -\frac{1}{n} \sum_{i=1}^{n} log \frac{e^{s\cdot(\cos{\theta_{y_i}} - m_2)}}{e^{s\cdot(\cos{\theta_{y_i}} - m_2)} + \sum_{j=1,j \neq y_i}^{C} e^{s \cdot \cos{\theta_{j}}}} , 
\end{align}

In such a margin scheme, algorithm is extremely easy and clear to implemet and has good convergence without tricky settings, and surface equation forms like
\begin{equation} \label{formu:am-softmax_surface}
s\cdot{(\cos{\theta_{y_i}} - m_2)} = s\cdot{\cos{\theta_{j}}}, 
\end{equation}

Cosine Metric indirectly reduces the angular distance by increasing the cosine value, but this effect will weaken as the cosine value increases, and the geometric interpretation is not clear. Instead, the angular distance can linearly represent the geodesic distance on the hypersphere. To this end, Arcface \cite{deng2019arcface} moves angular margin $ m $ to the inside of $ \cos(\theta_{y_i}) $ and adds it directly to the $ \theta_{y_i} $, 
\begin{equation} \label{formu:arcface}
L_{ArcS} =  -\frac{1}{n} \sum_{i=1}^{n} log \frac{e^{s\cdot\cos({\theta_{y_i}} + m_3)}}{e^{s\cdot\cos({\theta_{y_i}} + m_3)} + \sum_{j=1,j \neq y_i}^{c} e^{s \cdot \cos{\theta_{j}}}},
\end{equation}
where $ \theta \in \left[ 0, \pi - m_3 \right] $. In actual use, $ \cos({\theta_{y_i}} + m_3) =  \cos({\theta_{y_i}})\cos(m_3) - \sin({\theta_{y_i}})\sin(m_3) $, a relatively simple equivalent substitution. Similarly, surface equation is in the form as
\begin{equation} \label{formu:arcface_surface}
s\cdot{\cos{(\theta_{y_i} + m_3)}} = s\cdot{\cos{(\theta_{j})}}, 
\end{equation}

From original Softmax Loss (Eq.(\ref{formu:softmax})) to Normface (Eq.(\ref{formu:normface})), it is from optimizing inner-product similarity to optimizing cosine similarity (i.e. angle). And we can draw conclusions from Table \ref{tab:decision_boundaries}, from Modified A-Softmax (Eq.(\ref{formu:ma-softmax})) to ArcFace (Eq.(\ref{formu:arcface})), margin scheme makes decision boundaries change from common to separated and gradually get far apart. Hence margin forms between separeted boudaries. 


\subsection{A Toy Example}
\label{sec:a_toy_example}
In this section, a toy example on CIFAR-10 \cite{krizhevsky2009learning} dataset is presented. We use a relatively small and plain network, Resnet-18 \cite{he2016deep}, to perform feature extraction, and reduce the output number of the last FC layer to 3 (i.e. the feature dimension is 3) for the ease of visualization. Note that 
\begin{equation} \label{formu:softmax_grad}
\frac{\partial L_{s}}{\partial z_{y_i}} = \frac{e^{z_{y_i}}}{ e^{z_{y_i}} + \sum_{j=1, j \neq y_i}^{c} e^{ z_{j}} } - 1 = P_{y_i} - 1
\end{equation}
where $ L_{s} $ refers to all classification-based loss functions mentioned in Section.\ref{sec:Begin_with_Softmax}. And we can see from Eq.(\ref{formu:softmax_grad}) that the feature distribution and the gradient distribution are equivalent. To better reveal the anisotropic problem, we calculate the gradient w.r.t. target logit, $ \frac{\partial L}{\partial Z_{y_i}} $ , and plot it on the 3-D hypersphere to illustrate the distribution, instead of directly visualizing the features. 

We train the network with four different loss function: normface \cite{wang2017normface}, modified a-softmax, am-softmax \cite{wang2018additive} and arcface \cite{deng2019arcface}. These loss functions follow the hyperparameter settings in Section \ref{sec:implementation_details}. Then the resulting 3-D feature and its gradient are plotted in Figure \ref{fig:distribution_illus}. 

\subsection{Comparison}
\label{sec:comparison}
According to Eq. (\ref{formu:ma-softmax_surface})(\ref{formu:am-softmax_surface})(\ref{formu:arcface_surface}), $ m $  makes the conditions for the decision surface equation more stringent, forcing the angle to reduce and the intra-class distribution to compress. And the value of $ m $ controls how far to push the decision boundary, determining the margin size. 

For comparison, multiplicative way is applied directly on angle by $ m_1\theta_{y_i} = \theta_{j} $, the angular margin size depends on $ \theta $. Additive angular way brings about a fixed angular margin size by $ \theta_{y_i} + m_3 = \theta_{j} $ . However, the decision boundary measures in cosine space, constant $ m_3 $ leads to different decision margins for different classes. Additive cosine margin works in cosion space by $ \cos{\theta_{y_i}} - m_2 = \cos{\theta_{j}} $. With a fixed cosine margin size $ m_2 $, the boundaries are equally pushed towards class centers, resulting in the same decision margins for all classes. 
As can be seen from Figure \ref{fig:distribution_illus}, in some directions, the class boundary is close to class center, while in other directions, the boundary loses the gradient for further optimization at a distance far away from the center. The dynamics caused by the angular margin have weakened the anisotropy of the original intra-class distribution to some extent, whereas the situation is much more serious in the cosine margin scheme.

\subsection{Motivation}
\label{sec:motivation}
So, how to develop an effective loss function to alleviate the anisotropic problem of intra-class gradient distribution during training? Inspired by the idea that 'for the same metric, the gradients caused by the same distance gap should be equal', we propose the 'gradient-enhancing term' to explicitly enhance the gradients where large distance gap corresponds to very few gradient, making the boundary moving further towards the class center and equalizing the intra-class distribution. Where 'gradient-enhancing term' plays a role, the margin size is further expanded and feaure discrimination is further enhanced.

\section{Proposed Approach}
\label{sec:proposed_approach}
In this section, we will introduce the proposed \textbf{\emph{IntraLoss}} in detail. Section \ref{sec:gradient_enhancing_term} will first describe the definition of the loss funtion. Then Section \ref{sec:geometric_interpretation} analyze the novel function from gradient perspective. Lastly, Section \ref{sec:supplement} will supplement some problems and solutions of proposed approach.  

\subsection{Gradient-Enhancing Term}
\label{sec:gradient_enhancing_term}
Let us reconsider the classification-based loss as a function of target logit $ z_{y_i} $ , which is extracting from FC2, and the $ z_{y_i} $ and the angle $ \theta_{y_i} $ are negatively correlated. To make $ z_{y_i} $ converge to Optimum point, we can get a maxout function formulates as follow, 
\begin{equation} \label{formu:fin}
	f(O_p, z_{y_i}) = max(\beta - z_{y_i}, 0)
\end{equation}
where $ \beta = O_p - \gamma $ represents the \textbf{lower-bound} of $ z_{y_i} $. During optimization, $ z_{y_i} $ will continuely increase until it just exceeds $ O_p $. Unfortunately, the $ O_p $ is just a numerical point, we can't expect the entire class to converge to one point in actual use. It is not only difficult to converge, but we think this is also an overfitting of the training set, so $ \gamma $ is introduced. $ \gamma $ is a relaxation parameter representing the intra-class margin, so that the convergence condition of $ z_{y_i} $ relaxes from a strict point to a circular distribution centered on $ O_p $(i.e. $ |O_p - z_{y_i}|=\gamma $). The smaller the $\gamma$, the stronger the constraint of $ f(O_p, z_{y_i}) $, and the more concentrated the intra-class distribution, but the corresponding traing difficulty will increase or even not converge. 

Eq.(\ref{formu:fin}) is intuitive, it will suffers from \textbf{fixed-optimization} problem in training. No matter how far away from $ \beta $, $ z_{y_i} $ will constantly get "-1" gradient, even it is too large when near the boundary. And gradient decrease suddenly upon convergence.

Eq.(\ref{formu:softmax_grad}) has indicated that the gradient w.r.t. $ z_{y_i} $ is directly related to the target posterior probability $ P_{y_i} $ of input sample, and the value of gradient is in the range of $ \left[-1, 0\right] $. For network training, we hope that the enhanced gradient is dynamic to the optimization when maintaining the original value range to ensure the consistency and stability of the network convergence. 

To these ends, we propose the gradient-enhancing term as a modified \emph{Softplus} \cite{dugas2001incorporating} function, which is a differentiable approximation of maxout function. And it is a monotonically decreasing function of $ z_{y_i} $ and gradually decreases to zero as $ z_{y_i} $ increases. For the $ i $-th sample in the mini-batch, we compute the gradient-enhancing term as follows, 
\begin{equation} \label{formu:geti}
Get_{i} = \frac{1}{\alpha} \cdot log_e \left( e^{\alpha(\beta-z_{y_i})} + 1 \right), 
\end{equation}
where $ \frac{1}{\alpha} $ is for the convenience of derivation. The functional properties of $ \frac{\partial Get_i}{\partial z_{y_i}} $ is determined by $ \alpha $ and $ \beta $ together, details are in Section \ref{sec:geometric_interpretation}. Obviously, the gradient-enhancing term only works on target logit $ z_{y_i} $. 

\begin{figure*}  
	\centering
	\includegraphics[width=13cm]{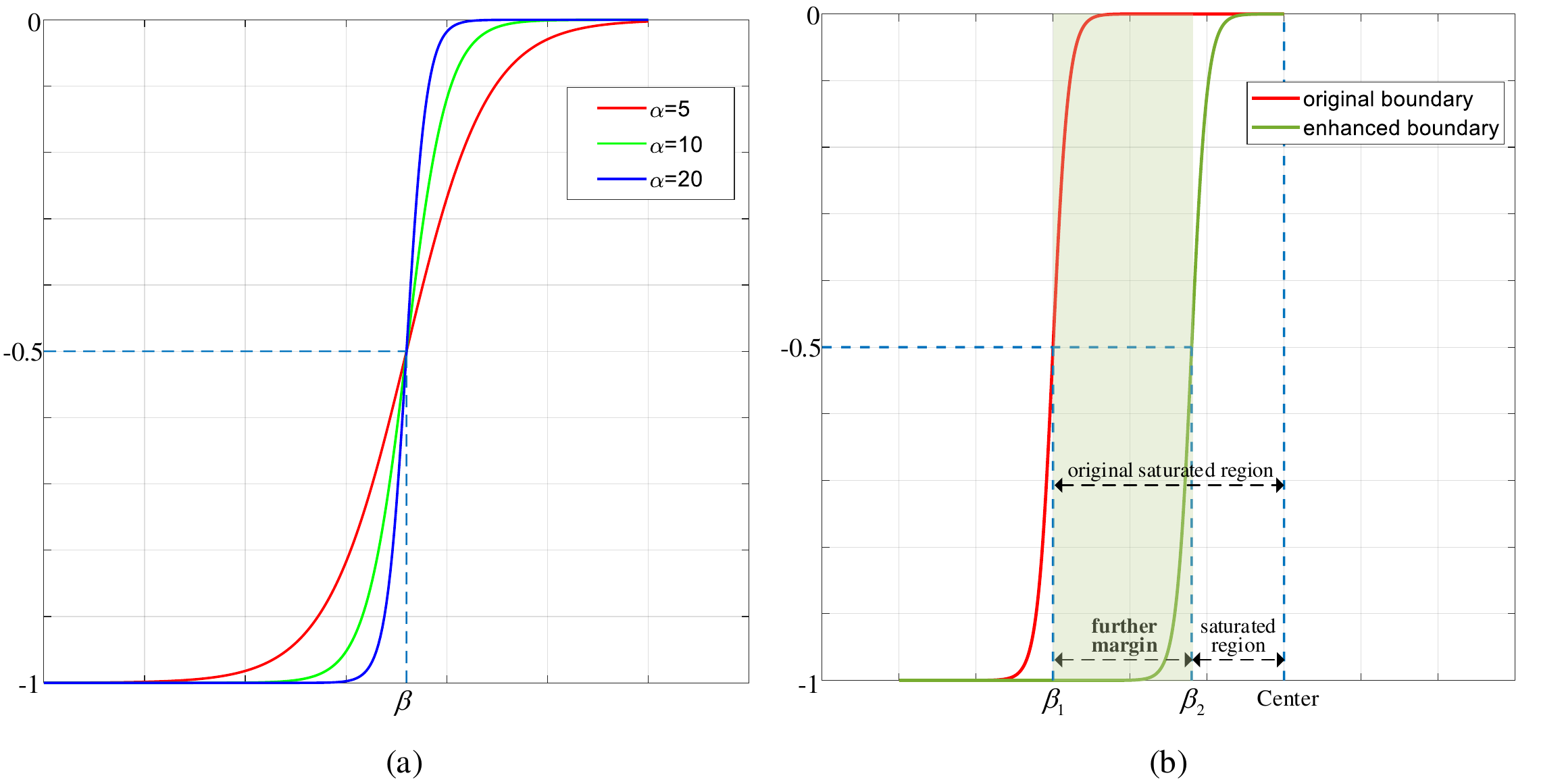}
	\caption{Illustration of the function. (a) The effect of $ \alpha $ and $ \beta $ on the function is shown: The larger the value of $ \alpha $, the steeper the rising part, and the narrower the gradient change area. (b) In the direction of incomplete optimization, the gradient brought by the gradient-enhancing term causes the calss boundary to continue to advance toward the class center, and the intra-class distribution continues to shrink to form a further margin. }
	\label{fig:gradfunc_illus}
\end{figure*}

\subsection{Adaptive Weighting}
\label{sec:adaptive_weighting}
Our objective is to design a loss function to perform adaptive gradient enhancement on different input samples and in different optimization stage. Here we introduce two weighting strategy to do the trick. 
\textbf{Self-paced Weighting.} To make it adaptive from sample to sample, the posterior similarity score of ground truth category $ P_{y_i} $ is adopted to measure the samples' optimization status. And the sample-wised weight $ (1 - P_{y_i}) $ is introduced to $ Get_i $, allowing each sample to learn its own pace. Not only that, such a weighting strategy makes training pay more attention to samples with poor optimization. With $ \gamma $ and $ (1 - P_{y_i}) $ the training process is more stable, and the convergence is also isotropic in all directions.

\textbf{Adaptive Unified Weighting.} Instead of balancing the each function with a fixed parameter $ \lambda $ as in previous work \cite{wen2016discriminative, zhao2019regularface}, we use $ w_{intra} $ to adaptively weight each sample batch, computed as below, 
\begin{equation}
	w_{intra} = \frac{1}{n} \sum_{i=1}^{n} P_{y_i}
\end{equation}
Not only that, $ w_{intra} $ also controls when $ L_{intra} $ participates in network optimization, just like an auto-switch: In the initial stage of training, the $ w_{intra} $ is close to 0, and the optimization is mainly based on $ L_{s} $. As $ L_{s} $ gradually converges and the inter-class margin is formed, the increased $ w_{intra} $ makes $ L_{intra} $ dominate and start to optimize the anisotropic intra-class distribution. At this time, $ L_{s} $ only serves to maintain the inter-class distribution. 

Finally, the \textbf{IntraLoss} is defined as, 
\begin{equation} \label{formu:intra_loss}
L_{intra} = w_{intra} \cdot \frac{1}{n} \sum_{i=1}^{n} (1 - P_{y_i}) \cdot Get_i, 
\end{equation}
where $ P_{y_i} = \frac{e^{z_{y_i}}}{ e^{z_{y_i}} + \sum_{j=1, j \neq y_i}^{c} e^{ z_{j}} } $. Then, we jointly supervise the network with classification-based loss function and Intraloss, the overall loss function is, 
\begin{align}
L_{all} &= L_{s} + L_{intra} 
\notag
\\
&= L_{s} + w_{intra} \cdot \frac{1}{n} \sum_{i=1}^{n} (1 - P_{y_i}) \cdot Get_i.
\end{align}

\subsection{Geometric Interpretation}
\label{sec:geometric_interpretation}
In this part, we analyze \emph{IntraLoss} by gradients, and we'll see how the gradient was enhanced. The gradient of $ L_{all} $ w.r.t. $ z_{y_i} $ is:
\begin{align} \label{formu:gradient}
\frac{ \partial L_{all} }{ \partial z_{y_i} } 
&= \frac{ \partial L_{s} }{ \partial z_{y_i} } + \frac{ \partial L_{intra} }{ \partial z_{y_i} } 
\notag
\\
&= \frac{ \partial L_{s} }{ \partial z_{y_i} } + w_{intra} \cdot (1 - P_{y_i}) \cdot \frac{ \partial Get_i }{ \partial z_{y_i} }
\notag
\\
&= \frac{ \partial L_{s} }{ \partial z_{y_i} } + w_{intra} \cdot (1 - P_{y_i}) \cdot \frac{-1}{ 1 + e^{-\alpha(\beta - z_{y_i})} }, 
\end{align}
where the $ \frac{ \partial L_{intra} }{ \partial z_{y_i} } $ is a weighted \textbf{sigmoid} function, ranging in [-1, 0]. Mathematically, $ \alpha $ controls the slope of the rising part of the sigmoid, and $ \beta $ is the '-0.5' gradient point. From Figure \ref{fig:gradfunc_illus}(a), we can see the effect of $ \alpha $ on the slope, and the slope directly determines how broad the gradient-changing area is.The gradient-changing area is what we call the class-boundary. Figure \ref{fig:gradfunc_illus}(b) shows that $ \beta $ is in the middle of changing area, so we can use $ \beta $ to locate the position of the boundary. 
In a conclusion, $ \alpha $ determines \textbf{"how"}, and $ \beta $ indicates \textbf{"where"}. 

Moreover, as discussed in Section \ref{sec:adaptive_weighting}, we consider $ w_{intra} $ to be an auto-switch to control the process of joint optimization. But the role of $ w_{intra} $ is more than that. From the perspective of gradient enhancement, $ w_{intra} $ can be regarded as a regional mask, which adaptively constraints the scope of $ \frac{ \partial L_{intra} }{ \partial z_{y_i} }  $ to corresponding intra-class distribution. Complete gradient enhancement is performed in the original saturated region of $ \frac{ \partial L_{s} }{ \partial z_{y_i} } $, and the enhancement gradient is gradually disappeared in the boundary region where $ P_{y_i} $ gradually reduces to 0. This mechanism prevents the leakage of the enhancement gradient from affecting other classes. 

\begin{figure*}  
	\centering
	\includegraphics[width=14cm]{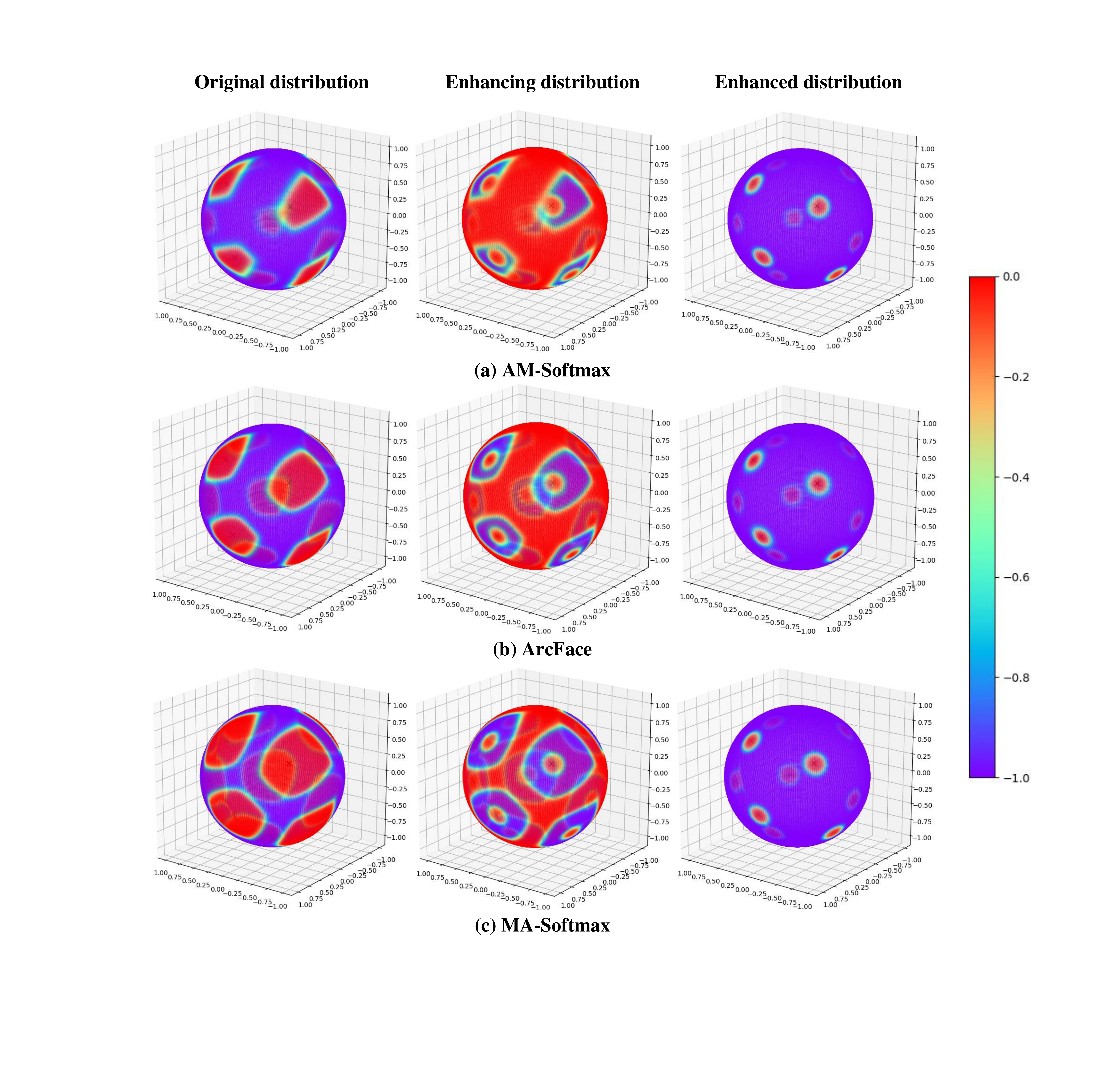}
	\caption{Illustration of gradient enhancing. It is easy to notice that the enhanced intra-class distribution is isotropoc, and the enhanced gradient is only related to the distance from the feature to center. In the anisotropic intra-class distribution, regions farther from the center will get more enhanced gradients (up to -1), and regions with shorter distances will get less gradients. And the saturation region of $ \frac{ \partial L_{intra} }{ \partial z_{y_i} } $ determines the new intra-class gradient boundary. }
	\label{fig:enhance_illus}
\end{figure*}

Our gradient-enhancing scheme has a clear geometric interpretation on the hypersphere manifold. As aforementioned, the gradient distribution of $ \frac{ \partial L_{s} }{ \partial z_{y_i} } $ is anisotropic. In Figure \ref{fig:gradfunc_illus}(b), we give a simple demonstration of the gradient-enhancing process in a single direction. In Figure \ref{fig:enhance_illus}, we show the changes in the intra-class distribution before and after gradient enhancement. 

\subsection{Supplementary details}
\label{sec:supplement}

%

\textbf{Parameter Setting.} \ There are three hyper-parameters for IntraLoss, i.e., $ \alpha, r, O_p $. $ \alpha $ determines the sharpness of boundary. A appropriate slope corresponds to the appropriate gradient change area, which can make the traing process more stable. $ \beta $ locates the position of the boundary. And $\gamma$ is a relaxation parameter, which represents the intra-class margin and determines the difficulty of convergence. We hope that IntraLoss can work stably with Softmax-based Loss functions, and the difficulty of convergence is moderate. For these reasons, we empirically set $ \alpha = 5 $ and $ \gamma = 0.9 $ in this paper. For $ O_p $, it depends on the margin scheme, and we show it in detail in Table \ref{tab:parameter_set}


\begin{table}
	\small
	\caption{Optimum $ O_p $ in different Margin Scheme. }
	\label{tab:parameter_set}
	\centering
	\setlength{\tabcolsep}{0.5cm}
	\begin{tabular}{l|c}
		\toprule[2pt]
		Margin Scheme                 & $ O_p $ \\
		\toprule[1pt]
		Multiplicative Angular Margin & s $\cdot$ cos(0 $\cdot$ $m_1$) \\
		Additive Cosine Margin        & s $\cdot$ (cos(0) - $m_2$) \\
		Additive Angular Margin       & s $\cdot$ cos(0 + $m_3$) \\
		
		\toprule[2pt]
	\end{tabular}
\end{table}

\textbf{Adaptive Two-stage Optimization.} \ $ L_{intra} $ focuses only on intra-class distributions, we don't want it to interfere with $ L_{s} $'s optimization process. So we divide the traing into two stages: In the first stage, $ L_{s} $ supervises the network only by itself. After the inter-class converges, $ L_{intra} $ joins the traing toperform gradient enhancement on the intra-class distribution to further learn discriminative features, and network gets better results under joint supervision of $ L_{intra} $ and $ L_{s} $. 

\textbf{Necessity of Joint Optimizing.} \ $ L_{s} $ traing the network alone will cause anisotropy problems in the intra-class distribution, but if we only supervise train CNNs by $ L_{intra} $ in second training stage, the inter-class distribution of features will be destroyed, and all features and centers will gradually cluster at one point, the fearures lose separability. For gradient enhancement of intra-class distribution, $ L_{s} $ is needed to maintain the original inter-class distribution. Simply using either of them will not get better results, so we need to combine them to train network together. 

\section{Experiments}
In this section, we first give the necessary inplementation details in Section \ref{sec:implementation_details} for the convenience of reproducing. Then we evaluate the proposed IntraLoss on three widely used face recognition benchmarks, including LFW \cite{huang2008labeled}, YTF \cite{wolf2011face} and CFP-FP \cite{cfp-paper}, and compare it with state-of-the-art classification-based loss functions to verify the effectiveness. 

\begin{figure*}  
	\centering
	\includegraphics[width=14cm]{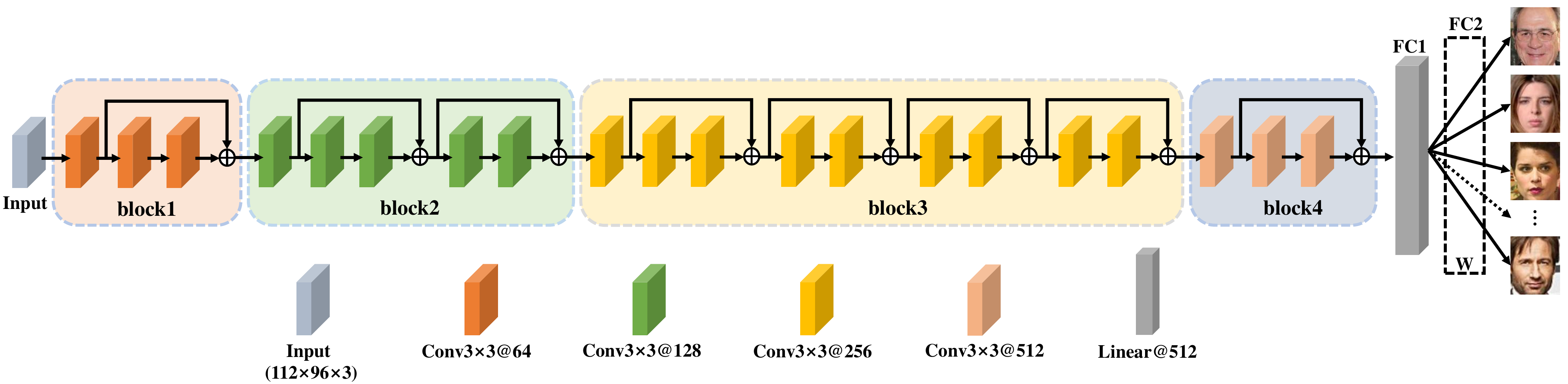}
	\caption{Illustration of the ResNet20 architecture. 'Conv3$ \times $3@X' represents the 3$\times$3 convolutional layer with X output channels. '$\bigoplus$' denotes element-wise sum operation, and W is the weight parameter of FC2. }
	\label{fig:network_arch}
\end{figure*}

\subsection{Implementation Details}
\label{sec:implementation_details}
\textbf{Preprocessing.} \ Following previous convention \cite{wen2016discriminative,liu2017sphereface}, only standard preprocessing is performed in this work. All the faces and landmarks in images are detected by MTCNN \cite{zhang2016joint}, and faces are aligned according to the five facial points (two eyes, nose and two mouth corners) by similarity transformation. Then the obtained faces are cropped and resized to be $ 112 \times 96 $. Finally, we also normalize each pixel (in [0, 255]) in RGB images by substracting 127.5 and then dividing by 128. 

\textbf{Training Data.} \ Publicly available dataset, CASIA-WebFace \cite{yi2014learning}, 
is 
used (after removing the overlapped identities between training set and test set) to train all the CNN models in the experiment. The dataset has 0.49M face images from 10575 identities.
And the training faces are horizontally flipped for data augmentation. Note that the scale of CASIA-WebFace is relatively small compared to other private datasets, such as VGGFace \cite{parkhi2015deep} (2M), DeepFace \cite{taigman2014deepface} (4M) and FaceNet \cite{schroff2015facenet} (200M).

\textbf{Network Settings.} \ It is widely recognized that deeper CNNs bring better performance. SphereFace \cite{liu2017sphereface} trains A-Softmax with different depth (4, 10, 20, 36 and 64) and the performance gradually increases as the depth deepens. Yet a deeper network requires more computing resources and longer training time. For fair comparison and the compromise between model performance and experimental efficiency, all compared methods in the experiments use the same architecture based on ResNet20 architecture, which is similar to \cite{liu2017sphereface} and has 20 convolutional layers based on residual unit \cite{he2016deep}. Figure \ref{fig:network_arch} gives the details that models accept $ 112 \times 96 $ RGB images and yields 512 dimensional feature at FC1. And loss layer (FC2) are appended after FC1 to perform classification during training. Pytorch \cite{paszke2019pytorch} is used to implement loss layers and CNNs. The models are trained from scratch with the batch size of 256 on one GTX 1080Ti GPU and updated through SGD algorithm with weight decay of 5e-4, and the momentum is set to 0.9. Due to the aditional gradient-enhancing phase, we extend the training process, so the learning rate is initialized to 0.1 and divided by 10 at 16k, 26k, 34k and 42k iterations. Finally, training procedure is finished at 46k iterations. 

\textbf{Evaluation Protocol.} \ Three benchmarks, LFW \cite{huang2008labeled}, YTF \cite{wolf2011face} and CFP-FP \cite{cfp-paper}, are utilized to evaluate the performance. We extract features from the output of FC1 (Figure \ref{fig:network_arch}) layer, and the original face features and its mirrored face features are concatenated to compose the final representation. Cosine similarity is performed as the measurement when comparing two face representations. On the LFW and YTF, standard ten-fold cross-validation are performed: The dataset is evenly divided into ten folds, nine of which are used to tune the best threshold and the accuracy is tested on the remaining fold. For the CFP-FP, we follow the official protocol.

\textbf{Different Loss Formulas.} \ As mentioned in Section.\ref{sec:comparison}, this anisotropy appears in both angular and cosine margin schemes, but is particularly prominent in cosine scheme. Hence it is reasonable to combine the proposed gradient-enhancing term $ L_{intra} $ with classification based large-margin scheme so as to learn large margin features while making intra-class distributions equally optimized in all directions. In this paper, we perform experiments with three loss function formulas: Normface + $ L_{intra} $ (Intra + Norm), Modified A-Softmax + $ L_{intra} $ (IntraLoss + MA), AM-Softmax + $ L_{intra} $ (IntraLoss + AM). For ArcFace, unfortunately, 
it is unstable in our implementation and is hard to compare.  
And the results shown in Table \ref{tab:performance_lfw}, \ref{tab:performance_ytf} and \ref{tab:performance_cfp} reveal the effectiveness of our proposed gradient-enhancing term. 

\textbf{Hyperparameter setting.} \ For the hyperparameter settings of the compared losses, we set them to fixed value and no longer discuss their impact on performance. Based on the previous works \cite{wang2017normface, ranjan2017l2}, we directly set scale parameter $ s $ to 30. And the margin parameter $ m $ of different schemes follows conresponding research: $ m_1 = 4 $ for multiplicative angular margin \cite{liu2017sphereface}, and $ m_2 = 0.35 $ for additive cosine margin \cite{wang2018cosface, wang2018additive}. 

\begin{table}
	\small
	\caption{Performance comparison (\%) with State-of-The-Art methods on LFW. \textbf{IntraLoss+[X]} means the joint supervision of gradient-enhancing term and loss function proposed in relevant paper, and $(\chi)$ indicates that this method conbines $\chi$ models. }
	\label{tab:performance_lfw}
	\centering
	\begin{tabular}{l|c|c}
		\toprule[2pt]
		Method                                   & Data & LFW   \\ 
		\toprule[2pt]
		DeepFace \cite{taigman2014deepface} (3)  & 4M &  97.35   \\
		FaceNet \cite{schroff2015facenet}        & 4M & 99.65     \\  
		DeepID2+ \cite{sun2015deeply} (25)       & 4M & 99.47  \\
		Center Loss \cite{wen2016discriminative} & 0.7M & 99.28  \\ 
		\hline
		Softmax Loss                             & \multirow{9}{*} { \shortstack{WebFace\\(0.49M)} } & 97.88 \\
		NormFace \cite{wang2017normface}         &                                                    & 98.22 \\ 
		SphereFace \cite{liu2017sphereface}      &                                                    & 98.88 \\ 
		MA-Softmax                               &                                                    & 99.10 \\  
		AM-Softmax \cite{wang2018additive}       &                                                    & 99.14 \\  
		ArcFace \cite{deng2019arcface}           &                                                    & 99.05 \\  
		IntraLoss+Norm                           &                                                    & 98.78 \\
		IntraLoss+MA                             &                                                    & 99.15 \\  
		\textbf{IntraLoss+AM}                    &                                           & \textbf{99.28} \\
		\toprule[2pt]
	\end{tabular}
\end{table}

\subsection{Experiments on LFW}
\label{sec:ex_lfw}
The LFW\cite{huang2008labeled} dataset consists of over 13000 web-collected face images from 5749 different identities, and only 1680 identities have two or more images. These images are captured in the wild conditions with large variations in pose, illumination, expression, color jittering and background, leading to \textbf{large intra-class variations}. Following the standard protocol of \emph{unrestricted with labeled outside data}, we evaluate our model on 6000 ground-truth matches(half of mathes are positive while the other half are negative) and report experiment results in Table \ref{tab:performance_lfw}. 

Table \ref{tab:performance_lfw} shows that, with the help of gradient-enhancing , our proposed \textbf{IntraLoss} method can improve the performance of all the softmax-based methods, among which the improvement of NormFace is the most significant(from 98.22\% to 98.78\%), and the best performance is obtained when combined with AM-Softmax(from 99.16\% to 99.28\%), in addition, methods MA-Softmax has also been improved from 99.10\% to 99.15\%.

\subsection{Experiments on YTF}
\label{sec:ex_ytf}
YTF\cite{wolf2011face} dataset includes 3,424 videos of 1,595 different person downloaded from YouTube, with the number of frames ranging from 48 frames to 6,070 frames. There are about 2.15 videos in YTF, and each video contains an average of 181.3 frames. We follow the evaluation protocpl similar to LFW on 5,000 video pairs, the only difference is that we adopt two methods to get feature of each video: using the feature averaging on all frames and using the feature of the middle frame of the corresponding video. 

As depicted in Table \ref{tab:performance_ytf}, conclusions similar to Table \ref{tab:performance_lfw} can be obtained from the result in "YTF(1)". When it comes to "YTF(avg)", we find something different: the overall results has improved by 1\% to 2\%, but the improvement brought by \textbf{IntraLoss} is not as prominent as before. 

\begin{table}
	\small
	\caption{Performance comparison (\%) with State-of-The-Art methods on YTF. YTF(1) means we represent each video with one frame, while YTF(avg) using all frames.}
	\label{tab:performance_ytf}
	\centering
	\begin{tabular}{l|c|c|c}
		\toprule[2pt]
		Method                                   & Data & YTF(1) & YTF(avg)  \\ 
		\toprule[2pt]
		DeepFace \cite{taigman2014deepface} (3)  & 4M & - & 91.4  \\
		FaceNet \cite{schroff2015facenet}        & 4M & - & 95.1    \\  
		DeepID2+ \cite{sun2015deeply} (25)       & 4M & - & 93.2 \\
		Center Loss \cite{wen2016discriminative} & 0.7M & - & 94.9  \\ 
		\hline
		Softmax Loss                             & \multirow{8}{*} { \shortstack{WebFace\\(0.49M)} }  & 89.54 & 91.94 \\
		NormFace \cite{wang2017normface}         &                                                    & 90.92 & 93.28 \\  
		MA-Softmax                               &                                                    & 92.59 & 94.73 \\  
		AM-Softmax \cite{wang2018additive}       &                                                    & 92.48 & 95.15 \\
		ArcFace \cite{deng2019arcface}           &                                                    & 92.66 & 95.01 \\  
		IntraLoss+Norm                           &                                                    & 91.30 & 93.58 \\
		IntraLoss+MA                             &                                                    & 92.62 & 94.75 \\  
		\textbf{IntraLoss+AM}                    &                                                    & \textbf{92.92} & \textbf{95.20} \\
		\toprule[2pt]
	\end{tabular}
\end{table}

\subsection{Discussion: LFW \& YTF} \label{sec:discuss_ly}
In Section \ref{sec:ex_lfw} and Section \ref{sec:ex_ytf}, we explored the results of IntraLoss on LFW and YTF, from which we draw the following two observations:

\begin{itemize}
	\item \textbf{Inconsistent performance improvement across strategies.} When there is only one image for verification of each identity, the performance improvement of IntraLoss is more prominent, and when the images used for verification increase, the improvement brought by our method will decay. Reminiscent of the large intra-class variance of the LFW and YTF datasets, we assume that our \textbf{IntraLoss is more suitable for dataset with large intra-class variance}. And the reason for the above phenomenon is the average strategy used in the YTF(avg). This strategy uses more data to get better results, and it reduces the intra-class variance to some extent, thereby weakening the role of our method. 
	\item \textbf{Inconsistent performance improvement across margin scheme.} Compared with the other two angular margin schemes, we found that always "Intra+AM"(cosine margin) can get better results, and during the experiment "Intra+AM" is more stable than "Intra+MA"(angular margin). "AM" imposes margin in cosine space, while margin scheme of "MA" works in angular space. Obviously, IntraLoss imposes constraints on $z_{y_i}$ in the consine space, so we assume that the cause of this phenomenon is the gap between the cosine space and angular space. Such a gap compromises the effect of our method on "MA". 
\end{itemize}

Anyway, the IntraLoss still improves the performance of Softmax-based loss function. 

\section{Experiments on CFP-FP} \label{sec:ex_cfp}
In Section \ref{sec:discuss_ly}, we find that IntraLoss may be more effective in large-variance data distribution. To further verify the effectiveness of our IntraLoss, we'll test it on CFP dataset\cite{cfp-paper} in this section. The dataset contains a total of 7000 images, composed of frontal and profile faces from 500 people, and each people has 10 frontal images and 4 profile images. 
\begin{table}
	\small
	\caption{Performance comparison with State-of-The-Art methods on CFP dataset(\%) .}
	\label{tab:performance_cfp}
	\centering
	\begin{tabular}{l|c|c}
		\toprule[2pt]
		Method                                   & Protocol & Ver.  \\ 
		\toprule[2pt]
		Softmax Loss                             & \multirow{9}{*} {\shortstack{FP}} & 85.51 \\  
		NormFace \cite{wang2017normface}         &          & 89.47 \\
		MA-Softmax                               &          & 93.10 \\
		AM-Softmax \cite{wang2018additive}       &          & 93.19 \\
		ArcFace \cite{deng2019arcface}           &          & 92.91 \\
		IntraLoss+NormFace                       &          & 93.16 \\
		\textbf{IntraLoss+MA}                    &          & \textbf{93.96} \\
		IntraLoss+AM                             &          & 93.90 \\
		\toprule[2pt]
	\end{tabular}
\end{table}

The test pipeline of CFP is similar to LFW and YTF, but the difference is that CFP performs frontal-profile(FP) face verification to test the algorithm in addition to the regular frontal-frontal(FP) face verification. In FP test protocol, algorithm suffers from extreme pose variance, besides other 'in the wild' variation. And the final results are reported in Table \ref{tab:performance_cfp}. 
 
From the results in Table \ref{tab:performance_cfp}, we see the effectiveness of IntraLoss in large-variance scenarios. In different large-margin methods, IntraLoss has achieved different degrees of improvement, and compared with the improvement in LFW and YTF, the improvement in CFP is more significant. The results prove that the gradient-enhancing method can not only further improve the existing large-margin method, but also is more suitable for large-variance problem.

\section{Conclusion}
\label{sec:conclusions}
In this paper, we propose an novel gradient-enhancing term, called IntraLoss, to guide softmax-based large-margin loss functions to learn face feature with smaller intra-class variation. During optimization, IntraLoss adaptively enhances the gradient of the area with anisotropy in the feature space, forcing the class boundary to contract further toward the center. Comprehensive experiments on several face benchmarks show the effectiveness of the method, and it is especially applicable to the large-variance problem.

{\small
\bibliographystyle{ieee_fullname}
\bibliography{intra-cas-refs}
}

\end{document}